\documentclass[11pt]{article}

\usepackage[final]{acl}

\usepackage{times}
\usepackage{latexsym}

\usepackage[T1]{fontenc}
\usepackage{amsmath}
\usepackage{amssymb}
\usepackage[utf8]{inputenc}

\usepackage{microtype}

\usepackage{inconsolata}

\usepackage{graphicx}

\usepackage{booktabs}
\usepackage{multirow}
\usepackage{graphicx}
\usepackage[table]{xcolor}
\usepackage{makecell}

\title{KEEPER: Learning Control-Sufficient Textual Memory \\for Long-Horizon LLM Agents}

\author{
  Jinkyeong Choi, Chaebin Jeong, Junsung Jeon and
  Donghyeon Park\thanks{Corresponding author.} \\
  Sejong University \\
  Seoul, Republic of Korea \\
  \texttt{jjinchoi@sju.ac.kr} \quad
  \texttt{chaibin2000@sju.ac.kr} \quad
  \texttt{jackjjs73@gmail.com} \quad
  \texttt{parkdh@sejong.ac.kr}
}

\begin{document}
\maketitle
\begin{abstract}
Recursive textual memory can carry earlier errors into later
updates, yet task performance alone does not establish memory
reliability.
We define self-consistency bias as the persistence of incorrect
claims already present in the incoming memory.
We present KEEPER, a comparative analysis of four auxiliary
objectives along information-content and update-sensitivity
axes within a shared recursive interface, examining how
supervision targets relate to agent performance.
Among evaluated conditions, reward-related prediction achieves
the highest task performance before and after offline
reinforcement learning, outperforming joint supervision.
A claim-level analysis of one baseline trajectory documents
error persistence despite corrective observations.
Together, these analyses motivate evaluating error persistence
and valid-information retention alongside task outcomes.
\end{abstract}

\section{Introduction}
\label{sec:introduction}

Long-horizon language agents need information from earlier
interactions to guide later decisions.
We use \emph{textual memory} to mean a natural-language summary
of past observations, actions, and task progress that an agent
uses for subsequent decisions.
In recursive memory systems, a generator updates this summary
using the previous memory and the latest
interaction~\cite{zhou2026mem1,yu2026memagent,zhen2026stephrl}.
We use \emph{self-consistency} to describe agreement between
successive memories on historical claims that new evidence
has not revised.
Preserving valid historical claims matters when the current
observation leaves earlier events unspecified, including
whether a prerequisite action has already been completed.

Agreement between successive memories does not establish
the factual correctness of the information they share.
Language models can repeat the same incorrect answer across
independent samples from a fixed input, and existing detection
methods struggle to identify such errors~\cite{tan2025consistent}.
Recursive memory adds a further route to agreement:
each update conditions on the previous memory, so a claim
can recur through reuse rather than independent regeneration.
If a memory incorrectly records a prerequisite as completed,
later updates may repeat that statement while new observations
leave the issue unresolved.
This concern is related to intrinsic self-correction:
in reasoning tasks, revising an answer without external
feedback does not reliably improve
performance~\cite{huang2024selfcorrect}.

We refer to the persistence of incorrect claims
already present in the incoming memory across
successive updates as \textbf{self-consistency bias}.
The focus is on the continuation of an existing error;
the generation of a new error is a separate event.
In agents that retrieve stored experiences, inaccuracies in
past experiences propagate to later tasks~\cite{xiong2026memory};
there, stored records are reused as retrieved, whereas recursive
memory is rewritten at every update.
Whether self-consistency bias occurs in recursively rewritten
memory, and under which update conditions, requires direct
evaluation in the memory system.

If an intervention reduces self-consistency bias,
that reduction alone is insufficient
to establish an improvement in memory quality.
An update that discards historical information can remove
errors while also losing correct facts needed for later decisions.
Improving recursive memory therefore requires reducing
self-consistency bias while preserving historical information
that remains correct and relevant to the task.

Existing approaches~\cite{zhou2026mem1,li2026mempo,wang2026selfsum}
train memory using task feedback,
memory-specific credit assignment, and learned summarization decisions.
These training signals can encourage the retention of useful
information.
However, improved task scores alone leave unclear how the
properties encouraged by training affect self-consistency bias
and the retention of valid historical information.
We therefore compare auxiliary objectives that target
different properties of recursive memory updates.

We organize the analysis along two axes.
The first concerns \emph{information content},
specifically which information
the memory generator's internal representation retains.
Approximate information-state theory motivates reward-related
and predictive information as complementary criteria under
explicit assumptions~\cite{subramanian2022ais,ni2024selfpredictive}.
Semantic agreement with reference memories provides a
comparison for these task-related signals.
The second concerns \emph{update sensitivity},
specifically how changes in the incoming memory affect
the subsequent update.
Recursive reuse motivates examining this sensitivity,
because errors in one memory become part of the input
to the next update.

We introduce \textbf{KEEPER} to analyze how auxiliary supervision
affects task outcomes within a fixed recursive memory interface.
We evaluate four auxiliary objectives, individually and jointly,
and compare task outcomes before and after offline reinforcement
learning.
For each objective, we specify a hypothesis linking its
supervision mechanism to recursive memory quality.
A complementary claim-level analysis examines how incorrect
claims persist across updates in one baseline trajectory.

We compare three objectives for \emph{information content}.
\textbf{(1) Reward Preservation}~(Sec.~\ref{sec:method:rp})
requires reward-related prediction, which we hypothesize
encourages representations to retain distinctions relevant
to task progress.
\textbf{(2) Predictive Preservation}~(Sec.~\ref{sec:method:pp})
requires successor prediction, motivating the retention
of future-relevant information beyond immediate reward.
\textbf{(3) Semantic Anchoring}~(Sec.~\ref{sec:method:sa})
aligns a learned projection with the semantic representation
of a reference memory.
We hypothesize that this alignment encourages the generator
to preserve valid historical information in its output.

\textbf{(4) Update Stability}~(Sec.~\ref{sec:method:us})
constrains sensitivity to incoming-memory perturbations,
motivating a reduction in their influence on subsequent updates.

We also evaluate all four objectives jointly.
The task results show that the benefits of individual objectives
do not necessarily accumulate under joint supervision.
Together with the error persistence observed in a baseline
trajectory, these findings motivate evaluating recursive memory
through both task outcomes and direct assessment of
self-consistency bias and valid-information retention.

Our contributions are as follows:
\begin{itemize}
    \item
    We compare four auxiliary objectives within a shared recursive
    memory interface, separating information content from update sensitivity.

    \item
    On ScienceWorld~\cite{wang2022scienceworld} and ALFWorld~\cite{shridhar2021alfworld}, Reward Preservation ranks highest
    among evaluated conditions before and after offline reinforcement
    learning, exceeding joint supervision.

    \item
    We document self-consistency bias despite corrective observations
    in one baseline trajectory, distinguishing persistence,
    recurrence after correction, and omission without explicit correction.
\end{itemize}
\section{Related Work}
\label{sec:related_work}

\subsection{Learning to Manage and Revisit Memory}
\label{sec:related_work:memory}

%
Memory-R1 learns memory operations through reinforcement
learning~\cite{yan-etal-2026-memory}, and AgeMem unifies
long-term and short-term memory management within the agent's
policy~\cite{yu-etal-2026-agentic}.
Self-Sum learns summarization timing and content~\cite{wang2026selfsum},
while MemPO assigns credit based on memory
effectiveness~\cite{li2026mempo}.
A-Mem links and updates memory records~\cite{xu2025amem};
ReMemR1 retrieves historical memories during updates for
long-context question answering~\cite{shi2026rememr1}, and ARC
uses reflection to revise context during information
seeking~\cite{yao2026arc}.
Using STEP-HRL's recursive progress-memory
interface~\cite{zhen2026stephrl}, we compare supervision targets
within a shared update structure.

\begin{figure*}[h]
  \centering
  \includegraphics[width=\textwidth]{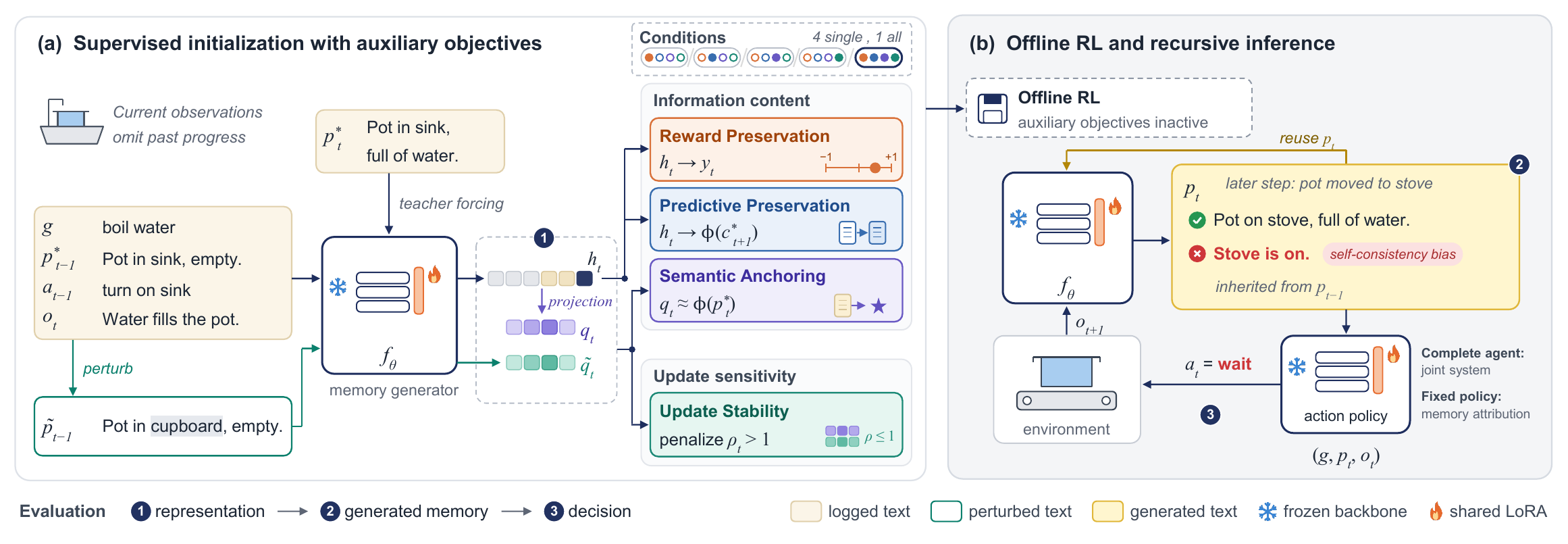}
  \caption{
Overview of auxiliary memory supervision and evaluation.
\textbf{(a)}~Supervised initialization: the memory generator $f_\theta$ encodes
the update context with the reference memory $p^{*}_{t}$ under teacher forcing.
Reward Preservation (RP), Predictive Preservation (PP), and Semantic Anchoring (SA)
supervise the information content of the representation $h_t$ or its projection
$q_t$ ($\phi$: fixed text encoder); Update Stability (US) penalizes the sensitivity
of $q_t$ to a perturbed incoming memory $\tilde{p}_{t-1}$.
We compare behavioral cloning (BC) alone, BC with each objective, and BC with all four.
\textbf{(b)}~Recursive inference, shared across conditions with auxiliary objectives inactive.
The generated memory keeps a valid fact but also an incorrect claim inherited from
$p_{t-1}$ (self-consistency bias), which misleads the policy; error persistence is
therefore evaluated together with valid-fact retention.
Numbered badges mark the three evaluation levels: representation, generated memory, and decision.
}
\label{fig:overview}
\end{figure*}

\subsection{Errors and Information Validity in Memory}
\label{sec:related_work:errors}

%
Prior work identifies consistent errors across independent
language-model samples~\cite{tan2025consistent}, error propagation
through retrieved experiences~\cite{xiong2026memory}, and reliance
on obsolete or invalidated memories in personalized
agents~\cite{uddin2026memora}.
MATCHA further shows that embedding-based similarity can
overlook contradictions~\cite{li2026matcha}.
We define \emph{self-consistency bias} as the persistence of
claims already incorrect in the incoming memory through
recursive rewriting.
Our claim-level analysis follows an existing incorrect claim,
distinguishing persistence, recurrence after correction, and
omission without explicit correction.

\subsection{Information Content and Update Sensitivity}
\label{sec:related_work:supervision}

%
Information-state theory and self-predictive RL motivate
reward and successor prediction as criteria for
decision-relevant representations under explicit
assumptions~\cite{subramanian2022ais,ni2024selfpredictive}.
The usefulness of self-predictive objectives also depends on
their relation to task values~\cite{pmlr-v258-khetarpal25a}.
Future-summary prediction~\cite{mahajan2026beyond} and
Semformer~\cite{yin2024semformer} use future-text and
response-semantic representations as auxiliary targets,
respectively.
These precedents motivate our comparison of reward-related
prediction, successor prediction, and reference-memory alignment.

%
Recurrent stability theory characterizes contractive
updates~\cite{miller2019stable}, while JacHess regularizes
representation derivatives with respect to input
embeddings~\cite{jukic2025robustness}.
We therefore compare information-content supervision with
penalties on sensitivity to incoming-memory perturbations,
asking whether their effects extend to generated memories and
reduce self-consistency bias while preserving valid historical
information.


\section{Methodology}
\label{sec:method}

%
%

Within \textbf{KEEPER}, we compare auxiliary objectives for
recursive textual memory and formulate hypotheses linking
their supervision mechanisms to self-consistency bias
and valid-information retention.
Fig.~\ref{fig:overview} summarizes the auxiliary
supervision during training and the shared recursive inference
procedure.
Our analysis separates \emph{information content}, which concerns
the distinctions encouraged in a representation, from
\emph{update sensitivity}, which concerns the response to changes
in the incoming memory.
Reward Preservation (RP) and Predictive Preservation (PP) target
information relevant to task outcomes and future contexts.
Semantic Anchoring (SA) provides a comparison based on
reference-memory content, while Update Stability (US) constrains
sensitivity to incoming-memory perturbations.

%

We distinguish three levels of evidence: the representational
property encouraged by supervision, its expression in generated
memory, and its consequences for subsequent decisions.
Improvement at one level provides a hypothesis to test at the
next level rather than establishing that effect.
For each objective, we derive a mechanism-specific expectation
and identify observations that would support the explanation.

\subsection{Recursive Memory and Supervision}
\label{sec:memory_principles}


Following STEP-HRL~\cite{zhen2026stephrl}, we consider an agent
executing a subtask $g$ with a textual memory $p_t$ at timestep $t$.
Given the previous memory $p_{t-1}$, previous action $a_{t-1}$,
and current observation $o_t$, the memory generator $f_\theta$,
with trainable parameters $\theta$, produces

{
\begin{equation}
p_t=f_\theta(g,p_{t-1},a_{t-1},o_t).
\label{eq:memory_interface}
\end{equation}
}
The action policy receives $g$, $o_t$, and the memory $p_t$ produced by Eq.~\eqref{eq:memory_interface}.
The generated memory supplies historical information beyond what the current observation specifies.


Behavioral cloning (BC) learns a policy through supervised imitation of demonstrations~\cite{foster2024behavior}.
We use BC for supervised initialization by minimizing
the negative log-likelihood of recorded reference outputs given their input contexts.


The auxiliary objectives supervise an internal representation
of the memory generator during training.
We use $\star$ to indicate reference text and $p_t^\star$
to denote the reference memory for the current update.
The representation $h_t$ is computed after the update context
and the complete reference memory under teacher forcing,
with reference output tokens supplied as context.
Let $\mathcal D$ denote the offline memory-update dataset;
$\mathbb E_{\mathcal D}$ denotes an average over its examples.
Appendix~\ref{app:representation} specifies the token position
used to extract $h_t$, the fixed text encoder, and the
prediction and projection functions.

\subsection{Information-Content Supervision}
\label{sec:information_content}



Information-content supervision specifies which historical
distinctions the memory representation should retain.
The following three objectives supervise these distinctions
through prediction of reward-related auxiliary labels,
prediction of successor contexts, and alignment with
reference-memory content.

\subsubsection{Reward Preservation (RP)}
\label{sec:method:rp}


Reward Preservation (RP) supervises information that distinguishes
reward-related outcomes.
Approximate information-state theory motivates expected-reward
prediction as a criterion for retaining decision-relevant
information~\cite{subramanian2022ais}, and latent-state modeling
uses reward prediction to constrain state
representations~\cite{gelada2019deepmdp}.
The relevant mechanism is the pressure to distinguish examples
with different targets: if their representations are
indistinguishable, a predictor receiving those representations
cannot account for the target difference.
When the difference depends on historical task progress,
preserving that information helps reduce prediction error.
Reward supervision therefore adds an explicit prediction
requirement beyond reference-token likelihood.


Let $y_t$ denote the reward-related target associated with
a logged update, and let $\widehat r$ predict $y_t$ from
the teacher-forced representation $h_t$.

We train $\widehat r$ using the prediction loss in
Eq.~\eqref{eq:reward_preservation}:
{
\begin{equation}
\mathcal L_{\mathrm{RP}}
=
\mathbb E_{\mathcal D}
\left[
\ell_{\mathrm{SL1}}\bigl(\widehat r(h_t),y_t\bigr)
\right],
\label{eq:reward_preservation}
\end{equation}
}

where $\ell_{\mathrm{SL1}}$ is the smooth-$L_1$ loss.
We expect this supervision to encourage distinctions that
explain the reward-related target.
The target $y_t$ is the reward label stored by STEP-HRL's
data converter, rather than the episode-level evaluation score.
Appendix~\ref{app:rp} details its construction, predictor warm-up,
and additional predictor training on detached low-level features.

%

We hypothesize that RP improves the preservation of task-progress
facts that help explain the reward-related target.
The mechanism predicts more accurate records of these facts
and reduced persistence of incorrect progress claims when
the distinction matters for target prediction.
Support would consist of improved reward-related prediction
accompanied by these claim-level changes.
The behavioral question concerns decisions that depend on
the recorded progress, such as whether a prerequisite remains
to be completed.

\subsubsection{Predictive Preservation (PP)}
\label{sec:method:pp}


Predictive Preservation supervises information that explains
subsequent contexts.
Two histories can have the same reward-related target while
leading to different continuations.
Reward prediction alone supplies no target difference that
requires those histories to remain distinguishable.
Approximate information-state theory therefore motivates
successor prediction alongside reward
prediction~\cite{subramanian2022ais,ni2024selfpredictive}.
Self-predictive representation learning also provides empirical
evidence for learning useful features through future
prediction~\cite{schwarzer2021spr,mahajan2026beyond}.
When successor targets differ, information that explains
the difference helps reduce the predictive loss, providing
additional supervision for distinctions that reward targets
leave unresolved.


Let $c_{t+1}^\star$ denote the recorded successor context,
and let $\phi$ be a fixed text encoder that produces
unit-normalized semantic embeddings.

A predictor $F$ maps $h_t$ to a prediction of the successor embedding. We train $F$ using the cosine loss in Eq.~\eqref{eq:predictive_preservation}:

{
\begin{equation}
\mathcal L_{\mathrm{PP}}
=
\mathbb E_{\mathcal D_{\mathrm{link}}}
\left[
1-\cos\bigl(F(h_t),\phi(c_{t+1}^\star)\bigr)
\right],
\label{eq:predictive_preservation}
\end{equation}
}

where $\mathcal D_{\mathrm{link}}$ contains examples with
a recorded successor context and $\cos$ denotes cosine similarity.
The expected effect is greater accessibility of historical
information useful for predicting logged continuations.
Appendix~\ref{app:pp} specifies the successor-matching rule,
the observation and memory fields in $c_{t+1}^{\star}$,
and the treatment of examples without a matched successor.

%

We hypothesize that PP preserves historical information that
distinguishes subsequent contexts even when reward-related
targets are similar.
Examples with comparable reward targets but different
continuations would provide a direct test of this hypothesis.
Improved successor prediction accompanied by better retention
of the facts that explain those continuations would support
the proposed mechanism.
The corresponding behavioral question is whether these
future-relevant distinctions help the policy select actions
whose consequences depend on them~\cite{pmlr-v258-khetarpal25a}.

\subsubsection{Semantic Anchoring (SA)}
\label{sec:method:sa}


Semantic Anchoring supervises correspondence with reference-memory
content.
Reward and predictive objectives select information according
to its usefulness for particular prediction targets.
Semantic alignment instead uses the reference memory itself
to define the representational target.
Matching distinct reference embeddings favors projected
representations that vary with reference content: a constant
projection cannot exactly match different embedding directions.
We therefore expect alignment to encourage content-dependent
representations to the extent that the fixed encoder
distinguishes the reference memories.
This condition provides a comparison between task-related
prediction and reference-based content supervision.


Let $q_t$ denote a learned, unit-normalized projection of $h_t$
into the embedding space of $\phi$.
We align $q_t$ with the reference-memory embedding using
the cosine loss in Eq.~\eqref{eq:semantic_anchoring}:
{
\begin{equation}
\mathcal L_{\mathrm{SA}}
=
\mathbb E_{\mathcal D}
\left[
1-\cos\bigl(q_t,\phi(p_t^\star)\bigr)
\right].
\label{eq:semantic_anchoring}
\end{equation}
}
The fixed encoder and normalized projection are specified in
Appendices~\ref{app:representation} and~\ref{app:sa}.

%

We hypothesize that SA helps generated memory retain correct,
task-relevant facts present in the reference memory.
Support would require improved alignment accompanied by better
retention of those facts, including factual distinctions that
broad semantic similarity can obscure~\cite{zha-etal-2023-alignscore}.
The reference output is already available to $h_t$ under
teacher forcing, so an alignment gain without corresponding
claim-level retention would be consistent with better extraction
of the supplied reference alone.
This comparison distinguishes reference-content retention
from improved representation matching.

\subsection{Update-Sensitivity Supervision}
\label{sec:update_sensitivity}


Update-sensitivity supervision concerns how a representation
changes when the incoming memory changes.
Holding the remaining update inputs fixed makes this response
the focus of the comparison.
US penalizes excessive sensitivity to sampled memory perturbations.

\subsubsection{Update Stability (US)}
\label{sec:method:us}


Update Stability constrains the influence of incoming-memory
perturbations on the next representation.
An erroneous claim in one memory becomes part of the input
to the next update, providing a route for its influence to persist.
Recurrent stability theory relates sensitivity to the
propagation of state differences: under a uniformly
contractive update and identical external inputs, differences
caused by an initial state decay over successive
updates~\cite{miller2019stable}.
This principle motivates penalizing sensitivity to perturbed
incoming memories~\cite{jukic2025robustness}.
The expected effect is a smaller influence of the sampled
perturbations on the subsequent representation.


We replace the reference incoming memory $p_{t-1}^\star$
with a perturbed memory $\widetilde p_{t-1}$ while holding
the subtask, action, observation, and reference output fixed.
Let $\widetilde q_t$ denote the resulting projected representation.
We define the sensitivity ratio $\rho_t$ and penalty as

{
\begin{equation}
\begin{aligned}
\rho_t
&=
\frac{\|q_t-\widetilde q_t\|_2}
{\|\phi(p_{t-1}^\star)-\phi(\widetilde p_{t-1})\|_2+\delta},\\
\mathcal L_{\mathrm{US}}
&=
\mathbb E_{\mathrm{valid}}
\left[\max(0,\rho_t-1)^2\right],
\end{aligned}
\label{eq:update_stability}
\end{equation}
}

where $\|\cdot\|_2$ is the Euclidean norm, $\delta>0$
stabilizes the denominator, and $\mathbb E_{\mathrm{valid}}$
averages over valid perturbation pairs.
The penalty in Eq.~\eqref{eq:update_stability} discourages
ratios above one in teacher-forced representations.
The contraction result motivates the objective without
establishing contraction of generated memories.
Appendix~\ref{app:us} specifies the perturbation operators,
valid-pair criteria, denominator stabilizer, and clean-branch
stop-gradient used to optimize the penalty.

We hypothesize that US reduces the persistence of errors
introduced through the incoming memory while retaining valid
historical facts.
Reduced sensitivity accompanied by lower self-consistency bias
and preserved facts would support this explanation.
If errors and valid facts both disappear, the result would
instead suggest reduced use of history.
Near-constant projections can also produce small sensitivity
ratios, so a lower penalty alone does not establish a more
reliable memory update.

\subsection{Comparison Conditions}
\label{sec:method:comparison}
\label{sec:training_inference}

We compare behavioral cloning (BC) alone, BC with each auxiliary objective individually, and BC with all four auxiliary objectives.
All conditions use the recursive memory and representation interfaces defined in Section~\ref{sec:memory_principles}.
The selected auxiliary terms determine the supervision condition:

{
\begin{equation}
\begin{aligned}
\mathcal L_{\mathrm{total}}
={}&\mathcal L_{\mathrm{BC}}
+\lambda_{\mathrm{RP}}\mathcal L_{\mathrm{RP}}
+\lambda_{\mathrm{PP}}\mathcal L_{\mathrm{PP}}\\
&+\lambda_{\mathrm{SA}}\mathcal L_{\mathrm{SA}}
+\lambda_{\mathrm{US}}\mathcal L_{\mathrm{US}}.
\end{aligned}
\label{eq:cstsa_objective}
\end{equation}
}

In Eq.~\eqref{eq:cstsa_objective}, $\mathcal L_{\mathrm{total}}$
combines behavioral cloning with the selected auxiliary
representation losses.
The behavioral cloning loss $\mathcal L_{\mathrm{BC}}$ is the
reference-output negative log-likelihood for the agent's
policy roles.
The nonnegative weights $\lambda_{\mathrm{RP}}$,
$\lambda_{\mathrm{PP}}$, $\lambda_{\mathrm{SA}}$, and
$\lambda_{\mathrm{US}}$ correspond to RP, PP, SA, and US, respectively~(Tab.~\ref{tab:auxiliary_weights}).
The baseline sets all four weights to zero, each individual
condition activates only its corresponding weight, and the
joint condition activates all four.

We also examine checkpoints obtained through subsequent
offline reinforcement learning following
STEP-HRL~\cite{zhen2026stephrl}.
The auxiliary objectives apply during supervised initialization
and remain inactive during offline reinforcement learning.
The auxiliary encoder and prediction and projection functions
are absent during execution.
Because training updates parameters shared by memory generation
and action selection, complete-agent task scores reflect
the jointly trained system.

Appendix~\ref{app:weights} specifies the auxiliary coefficients
for each condition, and Appendix~\ref{app:rp} describes
the reward predictor's additional head-only training.
Target construction, representation extraction, and perturbation
procedures are detailed in
Appendix~\ref{app:supervision_details}.


\begin{table*}[h]
\centering
\small
\setlength{\tabcolsep}{7pt}
\renewcommand{\arraystretch}{1.08}
\begin{tabular}{clcccc}
\toprule
& & \multicolumn{2}{c}{ScienceWorld}
& \multicolumn{2}{c}{ALFWorld} \\
\cmidrule(lr){3-4}\cmidrule(lr){5-6}
Variant & Initialization & Seen $\uparrow$ & Unseen $\uparrow$ & Seen $\uparrow$ & Unseen $\uparrow$ \\
\midrule
\multicolumn{6}{l}{\textit{Supervised initialization (BC)}} \\
\midrule
\textit{A} & BC
& 48.39 & 45.23 & 93.57 & 92.54 \\
\textit{B} & BC + RP~(Sec.~\ref{sec:method:rp})
& \textbf{49.39} & \textbf{49.01} & \textbf{96.43} & \textbf{95.52} \\
\textit{C} & BC + PP~(Sec.~\ref{sec:method:pp})
& 45.96 & 44.46 & 92.86 & 91.79 \\
\textit{D} & BC + SA~(Sec.~\ref{sec:method:sa})
& 41.85 & 40.90 & 92.14 & 90.30 \\
\textit{E} & BC + US~(Sec.~\ref{sec:method:us})
& \underline{49.12} & \underline{48.63} & \underline{95.71} & \underline{94.78} \\
\textit{F} & BC + RP + PP + SA + US
& 48.32 & 47.73 & 92.86 & 91.79 \\
\midrule
\multicolumn{6}{l}{\textit{After offline RL}} \\
\midrule
$\mathrm{A}^{\prime}$ & BC
& 45.46 & 42.11 & 94.29 & 92.54 \\
$\mathrm{B}^{\prime}$ & BC + RP
& \textbf{52.43} & \textbf{50.48} & \textbf{95.71} & \textbf{94.78} \\
$\mathrm{C}^{\prime}$ & BC + PP
& 47.81 & 46.31 & 93.57 & 92.54 \\
$\mathrm{D}^{\prime}$ & BC + SA
& 46.82 & 44.62 & 92.86 & 91.04 \\
$\mathrm{E}^{\prime}$ & BC + US
& 49.52 & 49.02 & \underline{95.00} & \underline{94.03} \\
$\mathrm{F}^{\prime}$ & BC + RP + PP + SA + US
& \underline{50.28} & \underline{49.65} & 93.57 & 93.28 \\
\bottomrule
\end{tabular}
\caption{
Comparison of auxiliary supervision before and after offline RL
on ScienceWorld and ALFWorld using Llama-3.2-3B-Instruct and
the STEP-HRL~\cite{zhen2026stephrl} protocol.
ScienceWorld reports average task scores, and ALFWorld reports
success rates (\%); higher values are better.
Variants A through F differ in their supervised objectives.
Primed variants apply the unchanged STEP-HRL offline RL procedure
to the corresponding initialized checkpoints, with auxiliary
objectives inactive.
BC: behavioral cloning; RP: Reward Preservation;
PP: Predictive Preservation; SA: Semantic Anchoring;
US: Update Stability.
The best and second-best scores in each column within each
training stage are marked in \textbf{bold} and
\underline{underlined}, respectively.
}
\label{tab:main_results}
\end{table*}

\section{Experiments}
\label{sec:experiments}

We examine how auxiliary supervision of recursive memory affects
whole-agent task scores on ScienceWorld and ALFWorld.
We compare each objective individually and all four jointly,
before and after offline reinforcement learning (RL).
A qualitative analysis of a recorded memory trajectory then
examines how incorrect claims persist, recur, and disappear
across updates.

\subsection{Experimental Setup}
\label{sec:experimental_setup}

\paragraph{Benchmarks and metrics.}
We use ScienceWorld~\cite{wang2022scienceworld} and
ALFWorld~\cite{shridhar2021alfworld}. 
ScienceWorld evaluates multi-step scientific reasoning in an
interactive text environment, while ALFWorld evaluates
language-conditioned household task execution.
We follow the data splits and evaluation protocol of
STEP-HRL~\cite{zhen2026stephrl}.
The ScienceWorld split includes 1,483 training task instances,
194 seen evaluation instances, and 211 unseen evaluation instances.
The ALFWorld split includes 3,211 training task instances,
140 seen evaluation instances, and 134 unseen evaluation instances.
We report average task scores on ScienceWorld and task success
rates on ALFWorld, separately for seen and unseen instances.
Higher values indicate better task completion.

\paragraph{Implementation details.}
We use Llama-3.2-3B-Instruct and follow the official
STEP-HRL implementation~\cite{zhen2026stephrl} for the
hierarchical agent architecture, demonstration data,
training and evaluation settings, and offline RL data construction.
Supervised initialization uses low-rank adaptation with AdamW,
a learning rate of $10^{-4}$, a total batch size of 128,
and five training epochs.
Each evaluation episode allows at most 50 environment steps.
Appendix~\ref{app:shared_protocol} summarizes the inherited protocol,
and Appendix~\ref{app:supervision_details} specifies the auxiliary
representations, targets, coefficients, and gradient routing.
The auxiliary objectives apply only during supervised initialization;
we retain the original offline RL procedure for the primed conditions
(Appendix~\ref{app:offline_rl}).
The auxiliary encoder and prediction and projection functions
are absent during execution.

\paragraph{Comparison conditions.}
We compare BC, BC + RP, BC + PP, BC + SA, BC + US, and
BC + RP + PP + SA + US, following
Section~\ref{sec:method:comparison}.
BC denotes supervised initialization of the STEP-HRL agent.
Each individual condition activates one auxiliary objective
in Eq.~\ref{eq:cstsa_objective}, while the joint condition
activates all four.
Variants A through F denote these initialization conditions;
primed variants denote the corresponding checkpoints after offline RL.

\subsection{Main Results}
\label{sec:main_results}

Tab.~\ref{tab:main_results} compares the supervision conditions
before and after offline RL.

\paragraph{Individual objectives.}
After supervised initialization, RP and US yield higher task
scores than BC on both seen and unseen splits of both benchmarks,
whereas PP and SA yield lower scores.
RP ranks highest across these comparisons.
The observed benefit therefore depends on the supervision signal.

\paragraph{Joint supervision.}
Applying all four objectives together yields lower task scores
than RP and US individually after supervised initialization.
The joint condition exceeds BC only on unseen ScienceWorld tasks.
Thus, the gains observed with individual objectives do not
accumulate under joint supervision in this setting.

\paragraph{Subsequent offline RL.}
RP remains the highest-scoring condition after offline RL,
but its ScienceWorld scores increase while its ALFWorld success
rates decrease relative to supervised initialization.
The joint condition improves across both splits of both benchmarks,
whereas BC scores decrease on ScienceWorld.
Offline RL therefore has different outcomes across initialization
conditions and benchmarks.

\paragraph{Scope of the comparison.}
These results compare complete agents: the memory generator
and controller share trainable parameters.
Higher task scores alone do not establish reduced self-consistency
bias or improved retention of valid historical information.
The following qualitative analysis examines observed memory
errors, without attributing their frequency to auxiliary supervision.

\begin{table}[h]
\centering
\small
\setlength{\tabcolsep}{3pt}
\renewcommand{\arraystretch}{1.08}
\begin{tabular}{@{}clcc@{}}
\toprule
Variant & Initialization & Seen $\uparrow$ & Unseen $\uparrow$ \\
\midrule
A & BC & 94.29 & 94.03 \\
B & BC + RP & 96.43 & 96.27 \\
C & BC + PP & 95.00 & 94.78 \\
D & BC + SA & 92.14 & 89.55 \\
E & BC + US & 95.71 & 94.03 \\
F & BC + RP + PP + SA + US & 93.57 & 93.28 \\
\bottomrule
\end{tabular}
\caption{
Comparison of auxiliary supervision using Qwen3.5-2B
on ALFWorld with the STEP-HRL~\cite{zhen2026stephrl} protocol.
The checkpoint is \texttt{Qwen/Qwen3.5-2B}.
All six conditions use supervised initialization without offline RL.
We report success rates (\%) on 140 seen and 134 unseen episodes;
higher values are better.
Variant definitions match Tab.~\ref{tab:main_results}.
}
\label{tab:qwen_alfworld}
\end{table}

\subsection{Generalization Across Backbones}
\label{sec:backbone_generalization}

Tab.~\ref{tab:qwen_alfworld} compares the same supervision
conditions using Qwen3.5-2B on ALFWorld after supervised
initialization, without subsequent offline RL.

\subsection{Analysis of Recursive Memory}
\label{sec:memory_analysis}

We examine a trajectory selected post hoc by comparing
claims about pot contents and agent location with recorded
environment observations.
Both figures describe the same episode, first within a single
update and then across successive updates.
Appendix~\ref{app:case_protocol} defines the claim categories
and distinguishes persistence, recurrence, and omission.

\begin{figure}[h]
  \centering
  \includegraphics[width=\linewidth]{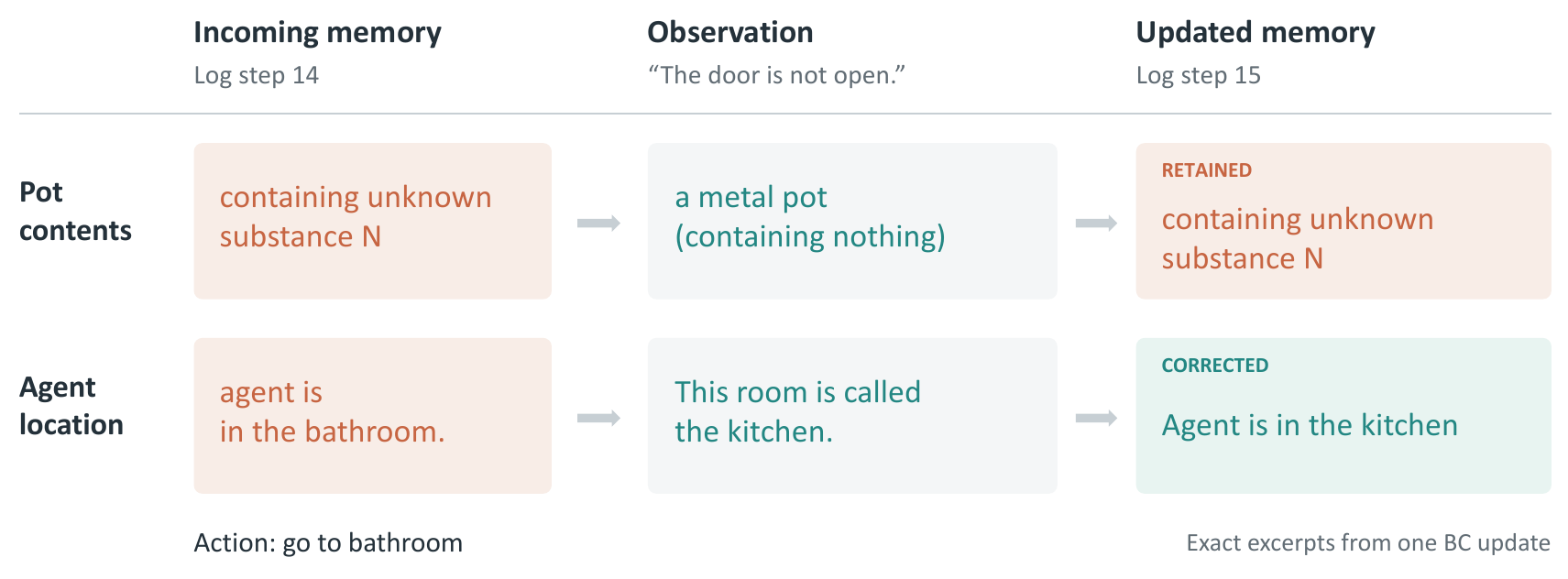}
  \caption{
Different responses to corrective evidence within one memory update.
The observation contradicts both the incoming location claim
and the claim about pot contents.
The updated memory corrects the location while retaining
the incorrect claim about pot contents.
The figure presents excerpts from the selected trajectory,
with line breaks added for readability.
}
  \label{fig:memory_update_case}
\end{figure}

\paragraph{Different responses within the same update.}
Fig.~\ref{fig:memory_update_case} shows an incoming memory
that places the agent in the bathroom and describes a metal pot
as containing an unknown substance.
The observation identifies the kitchen and reports an empty pot.
The updated memory corrects the location but retains the
incorrect claim about the pot contents.
An update can therefore incorporate corrective evidence for
one claim while preserving another that the observation contradicts.

\begin{figure}[h]
  \centering
  \includegraphics[width=\linewidth]{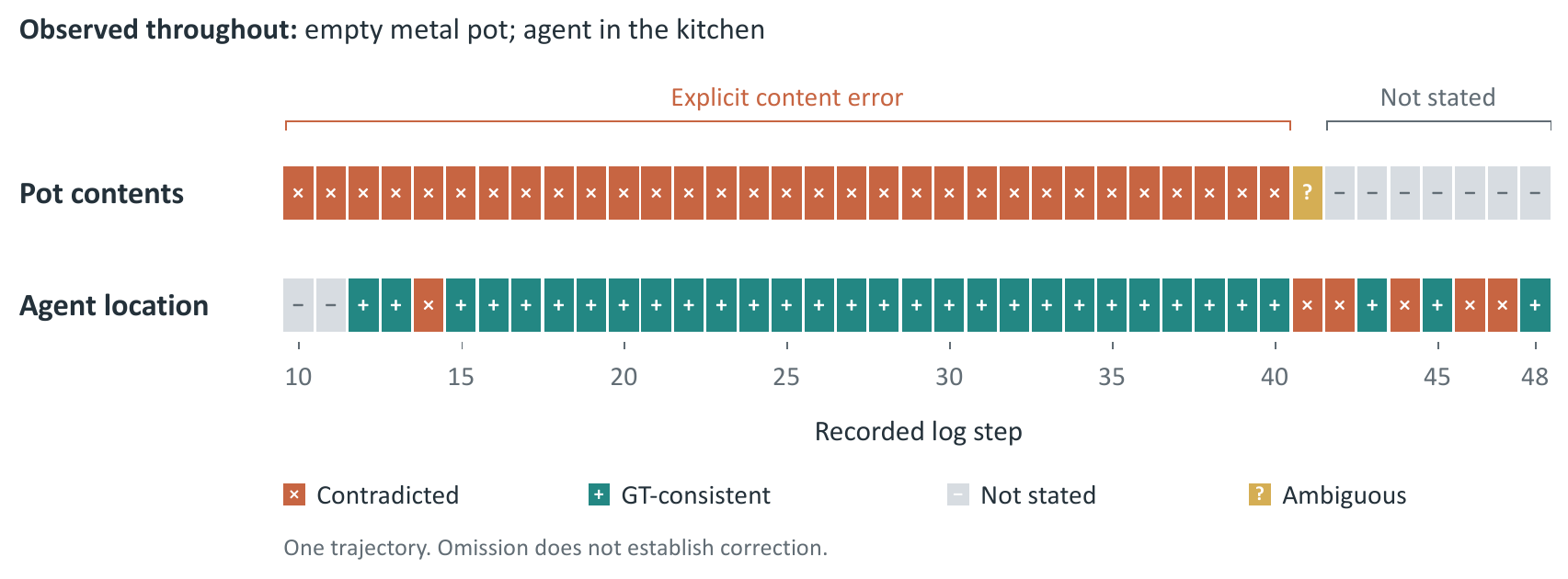}
  \caption{
Claim trajectories in the same episode as
Fig.~\ref{fig:memory_update_case}.
The recorded environment state remains unchanged: the metal pot
is empty and the agent is in the kitchen.
Symbols distinguish contradicted claims ($\times$), claims
consistent with the environment state ($+$), omitted claims ($-$),
and ambiguous claims ($?$).
The incorrect claim about pot contents persists before becoming
ambiguous and then omitted, while the location claim becomes
incorrect again after correction.
}
  \label{fig:memory_claim_trajectory}
\end{figure}

\paragraph{Different trajectories across updates.}
Fig.~\ref{fig:memory_claim_trajectory} follows the same claims
through the recorded chain.
The incorrect claim about pot contents persists across successive
updates despite repeated observations of an empty pot.
The location claim becomes correct and later becomes incorrect
again, distinguishing error recurrence from uninterrupted persistence.
The claim about pot contents eventually becomes ambiguous and
then disappears, without an explicit correction.

\paragraph{Persistence, recurrence, and omission.}
The persistent claim about pot contents illustrates
self-consistency bias despite corrective evidence.
The location claim shows that an error can recur after
correction, while the eventual omission of the claim about
pot contents does not demonstrate correction.
These observations distinguish persistence, recurrence,
and omission within the selected trajectory.
\section{Conclusion}

Our KEEPER has provided a comparative analysis of how four auxiliary
supervision objectives affect task outcomes within a shared
recursive memory interface, distinguishing information content
from update sensitivity.
On ScienceWorld and ALFWorld, Reward Preservation and Update
Stability have improved over behavioral cloning after supervised
initialization, whereas Predictive Preservation and Semantic
Anchoring have yielded lower task scores and success rates.
Reward Preservation has ranked highest both before and after
offline reinforcement learning, exceeding joint supervision.
The effects of subsequent offline reinforcement learning have
varied across benchmarks and initialization conditions.
A complementary baseline analysis has documented error
persistence despite corrective observations.
Together, these findings have shown that the task-level benefits
of auxiliary memory supervision vary with the supervision target
and training condition, and that the benefits of individual
objectives do not necessarily accumulate under joint supervision.

\section*{Limitations}

Our evaluation covers two language-model backbones within
one agent architecture, with the second backbone evaluated
only on ALFWorld.

Shared trainable parameters prevent attributing task-score
differences solely to memory generation.
Teacher-forced representation supervision does not establish
generated-memory correctness.
The sensitivity penalty evaluates projected representations
under sampled text perturbations and does not guarantee
global contraction.

Our qualitative evidence comes from one baseline trajectory
selected post hoc.
It does not establish error prevalence or the effects of
auxiliary supervision on error persistence and valid-information
retention.

\bibliography{custom}

\appendix

\section{Appendix}
\label{sec:appendix}

\subsection{Shared Experimental Protocol}
\label{app:shared_protocol}

Except for the auxiliary supervision described in
Appendix~\ref{app:supervision_details}, we follow the official
STEP-HRL implementation~\cite{zhen2026stephrl} for the hierarchical
agent, demonstration data, observation processing, policy prompts,
training, offline RL data construction, and evaluation.
The main comparison uses \path{meta-llama/Llama-3.2-3B-Instruct}.
The high-level policy, low-level policy, and memory generator
share the language-model backbone and its trainable LoRA adapter.
The auxiliary functions operate during supervised initialization
and are absent during environment interaction.

\subsubsection{Training and Evaluation}
\label{app:training_protocol}

Table~\ref{tab:shared_settings} summarizes the inherited settings.
Supervised initialization uses reference-output likelihoods for
all three policy roles. Following the implementation, the
high-level loss is evaluated every fourth training minibatch;
the low-level and memory-generation losses are evaluated at
every minibatch. Auxiliary losses use the memory-update batch.
The auxiliary functions use the same learning rate as the
trainable backbone parameters unless a separate head learning
rate is specified.

\begin{table}[htbp]
\centering
\small
\setlength{\tabcolsep}{3pt}
\begin{tabular}{@{}p{0.55\linewidth}p{0.40\linewidth}@{}}
\toprule
Setting & Value \\
\midrule
Optimizer & AdamW \\
Adam coefficients & $(0.9,0.999)$ \\
Optimizer epsilon & $10^{-8}$ \\
Weight decay & $0.01$ \\
Gradient clipping & $1.0$ \\
Model precision & bfloat16 \\
LoRA rank / scaling & $16$ / $32$ \\
LoRA dropout & $0.05$ \\
LoRA targets & Query, key, value, output projections \\
BC learning rate & $10^{-4}$ \\
BC epochs & $5$ \\
Effective batch size & $128$ \\
BC / evaluation seed & $0$ \\
Environment steps per episode & $50$ \\
Evaluation action-token budget & $32$ \\
Evaluation memory-token budget & $150$ \\
Evaluation decoding & Greedy \\
\bottomrule
\end{tabular}
\caption{Settings inherited from the official STEP-HRL implementation.
The batch size is the effective batch size in the released
DeepSpeed configuration. The model identifier follows the main comparison.}
\label{tab:shared_settings}
\end{table}

The ScienceWorld split contains 1,483 training instances,
194 seen evaluation instances, and 211 unseen evaluation instances.
The ALFWorld split contains 3,211 training instances,
140 seen evaluation instances, and 134 unseen evaluation instances.
ScienceWorld reports the average task score, including partial
completion, whereas ALFWorld reports the percentage of successful
episodes. We retain the upstream observation processing and
full-split evaluation for the reported comparisons.

\subsubsection{Subsequent Offline RL}
\label{app:offline_rl}

The primed conditions initialize the STEP-HRL offline RL stage
from their corresponding supervised checkpoints.
We retain the upstream construction of expert and collected
transition data and the original actor and critic updates.
All four auxiliary coefficients are zero during this stage,
and the critics retain the STEP-HRL architecture.
Thus, the RL comparison concerns subsequent optimization of
different initializations under the same procedure.

The released configuration uses three RL epochs, an actor
learning rate of $10^{-5}$, a critic learning rate of $10^{-4}$,
and a critic warm-up of 100 training iterations.
The discount factor is $0.99$, the target-network update
coefficient is $0.2$, and the expectile parameter is $0.95$.
The RL configuration points to the same DeepSpeed batch
configuration as BC. The dataset loader appends the configured
collected files to the expert data for each policy role.
We follow this implementation rather than imposing a separate
mixture ratio in the auxiliary-supervision code.

\subsection{Auxiliary Supervision Details}
\label{app:supervision_details}

This section specifies the representations, auxiliary functions,
target construction, and gradient routing used by the supplied
implementation. The implementation defaults are distinguished
from the condition-specific activation of each objective.

\subsubsection{Representations and Auxiliary Functions}
\label{app:representation}

For each memory update, the input contains the subtask,
previous reference memory, executed action, and resulting
observation. We append the reference output memory and an
end-of-sequence token, following STEP-HRL tokenization.
The representation $h_t$ is the final-layer hidden state at
this end-of-sequence token, selected by the target-end mask.
Consequently, $h_t$ conditions on both the update context and
the reference output under teacher forcing.

The fixed text encoder $\phi$ defaults to the MiniLM checkpoint
\path{sentence-transformers/all-MiniLM-L6-v2}.
It remains in evaluation mode with frozen parameters and
produces unit-normalized embeddings of dimension $d=384$.
Let $H$ denote the backbone hidden dimension.
Table~\ref{tab:auxiliary_functions} specifies the trainable
functions applied to $h_t$.
The reward predictor produces a scalar, while the successor
predictor and semantic projection produce normalized vectors
in the encoder space.

\begin{table}[htbp]
\centering
\small
\setlength{\tabcolsep}{3pt}
\begin{tabular}{@{}ll@{}}
\toprule
Function & Architecture \\
\midrule
Reward predictor $\widehat r$ & $H\rightarrow\lfloor H/4\rfloor\rightarrow1$ \\
Successor predictor $F$ & $H\rightarrow\lfloor H/2\rfloor\rightarrow d$ \\
Projection for $q_t$ & $H\rightarrow d$ \\
\bottomrule
\end{tabular}
\caption{Trainable auxiliary functions. Each arrow denotes a linear
layer with a bias; hidden layers use ReLU.
$H$ is the backbone hidden dimension and $d=384$ is the default
text-embedding dimension.}
\label{tab:auxiliary_functions}
\end{table}

\subsubsection{Reward Preservation}
\label{app:rp}

RP uses the stored \texttt{reward} field of each memory-update
example as its target $y_t$.
This field follows STEP-HRL's converted training labels rather
than a newly annotated measure of memory quality.
In the upstream conversion, a subtask-final action receives
$+1$, other actions receive $0$, and the designated invalid-action
response receives $-1$; the converter's terminal handling is
retained. The memory-update example inherits the corresponding
action label. These auxiliary labels are distinct from the
benchmark's episode-level evaluation score.

The reward predictor uses smooth-$L_1$ loss with its default
transition parameter of one.
With the default warm-up of 500 training minibatches, the
progress-side reward loss initially updates the predictor using detached features.
After this warm-up, the loss also propagates through $h_t$ to
the shared trainable backbone parameters.
The warm-up counter advances once per minibatch, before accounting
for gradient accumulation.

When RP is active and low-level reward labels are available,
the same reward predictor also receives supervision at the
end of the reference action output.
The low-level features are detached by default, so this
additional term trains the shared predictor without sending
a direct gradient through the low-level representation.
Both reward terms use the RP coefficient.
The auxiliary reward supervision therefore includes a
predictor-training component as well as the progress-side
representation loss described in the main text.

\subsubsection{Predictive Preservation}
\label{app:pp}

PP uses a recorded successor target constructed from adjacent
rows of each stored memory-update dataset.
A row is linked to the next row when its reference output
appears exactly as the next row's previous-memory field.
For a linked pair, the primary successor target $c_{t+1}^{\star}$
concatenates the next row's resulting observation and reference
output memory, using observation and local-progress field labels.
The target contains these two fields; it does not append the
subtask or a future action.

Rows without a matched successor receive an empty target and
are excluded from the PP average.
The implementation prefers the \texttt{next\_state} column.
For legacy files without that column, it can use the
\texttt{next\_progress} column, which contains only the next memory.
These target definitions are different and must be kept consistent
across compared PP conditions.
The predictor minimizes cosine distance to the frozen embedding
of each nonempty target.
If a minibatch contains no eligible targets, it contributes
no PP loss.

\subsubsection{Semantic Anchoring}
\label{app:sa}

SA projects $h_t$ into the fixed encoder space and normalizes
the result to obtain $q_t$.
It minimizes cosine distance between $q_t$ and
$\phi(p_t^{\star})$, where $p_t^{\star}$ is the reference output
memory. The encoder target receives no gradient.
The projection also supplies the representation used by US
when that objective is active.
An improvement in this teacher-forced alignment objective
concerns the supplied reference representation; generated-memory
factuality is a separate property.

\subsubsection{Update Stability}
\label{app:us}

US replaces only the previous-memory field of the update input.
The subtask, action, resulting observation, and teacher-forced
reference output remain fixed.
The implementation shuffles the configured perturbation
operators and accepts the first operator that changes the text.
The default operators are:
\begin{itemize}
\item \textbf{Deletion:} remove one sentence or tagged memory unit
when at least two units are available.
\item \textbf{Substitution:} replace one object or location mention
with a different vocabulary entry of the same type.
\item \textbf{Injection:} insert a template-generated statement
about an object, location, or completed action.
\end{itemize}
The perturbation sampler uses a local random generator initialized
with the training seed plus 1,234.
A comma-separated clause deletion serves as a fallback when
the configured operators fail.
Empty previous memories and examples for which no operator
changes the text are excluded from the US average.
Substitution and injection require a vocabulary with both
object and location entries. Without that vocabulary, only
applicable deletion operations remain.
The perturbations are text edits, without an external factuality
judge; an injected statement is therefore not guaranteed to
contradict the environment.

The denominator uses the Euclidean distance between the frozen,
normalized embeddings of the clean and perturbed previous memories,
with stabilizer $\delta=10^{-4}$.
The numerator uses the distance between the corresponding
normalized projections of the teacher-forced hidden states.
The penalty is the mean squared hinge above a ratio of one.
By default, the clean projection is detached, and gradients
from US flow through the perturbed branch.
This routing preserves the numerical form of the sensitivity
ratio while making its optimization asymmetric.

The ratio measures sensitivity of the learned projection on
sampled pairs, rather than a global Lipschitz constant of
decoded memory. In the US-only condition, SA is inactive,
so reference-content alignment does not constrain this projection.
A nearly constant projection can produce a small penalty;
the penalty alone therefore does not establish factual correction
or retention of useful history.

\subsubsection{Condition Weights and Training Scope}
\label{app:weights}

Table~\ref{tab:auxiliary_weights} maps the six comparison
conditions to the implementation's default active coefficients.
Each individual condition sets the other three coefficients
to zero. The joint condition activates all four.
The code defaults are $0.5$ for RP and $1.0$ for PP, SA, and US;
an omitted configuration key retains its default value.

\begin{table}[htbp]
\centering
\small
\setlength{\tabcolsep}{4pt}
\begin{tabular}{@{}lcccc@{}}
\toprule
Condition & $\lambda_{\mathrm{RP}}$ & $\lambda_{\mathrm{PP}}$ & $\lambda_{\mathrm{SA}}$ & $\lambda_{\mathrm{US}}$ \\
\midrule
BC & 0 & 0 & 0 & 0 \\
BC + RP & 0.5 & 0 & 0 & 0 \\
BC + PP & 0 & 1 & 0 & 0 \\
BC + SA & 0 & 0 & 1 & 0 \\
BC + US & 0 & 0 & 0 & 1 \\
BC + all four & 0.5 & 1 & 1 & 1 \\
\bottomrule
\end{tabular}
\caption{Reference coefficient settings using the implementation
defaults for active terms. Columns follow the objective names
in the main text. All coefficients are zero during subsequent offline RL.}
\label{tab:auxiliary_weights}
\end{table}

The default shared-adapter mode allows auxiliary gradients
to update parameters also used by action selection.
The optional separate-adapter mode, spectral critics, and
additional diagnostic probes are outside the reported protocol.
The training implementation saves auxiliary-function parameters
separately from the policy adapter; environment interaction
uses the policy adapter without these auxiliary functions.

\subsection{Reading the Qualitative Memory Case}
\label{app:case_protocol}

The two memory-analysis figures in the main text describe
one recorded trajectory labeled BC baseline.
The update-level view aligns the previous memory, the
resulting observation, and the updated memory.
The trajectory-level view follows the same two claims about
pot contents and agent location over subsequent recorded steps.
Their reference is the environment observation, rather than
agreement between successive generated memories.

A claim is \emph{contradicted} when its stated content conflicts
with the recorded environment state, and \emph{consistent} when
it agrees with that state.
An \emph{omitted} claim is absent from the memory text.
An \emph{ambiguous} claim lacks enough detail for either
correctness judgment.
These categories describe the displayed case; they do not
constitute an aggregate automatic metric.
Persistence denotes an incorrect claim continuing across
successive updates. Recurrence denotes an incorrect claim
appearing again after a correct statement.
Omission alone provides no evidence that the generator has
replaced an incorrect claim with the correct one.




\end{document}